\documentclass[preprint,12pt]{elsarticle}

\usepackage{amssymb}
\usepackage{graphicx}
\usepackage{amsthm}
\usepackage[section]{placeins}
\usepackage{subfigure}
\usepackage{caption} 
\usepackage{booktabs} 
\usepackage{longtable}
\usepackage{multirow} 
\usepackage{array} 
\usepackage{mathtools}

\usepackage{lineno}

\journal{Heliyon}

\begin{document}

\begin{frontmatter}



\title{A NEW METHOD IN FACIAL REGISTRATION IN CLINICS BASED ON STRUCTURE LIGHT IMAGES}


\author[inst1]{Pengfei Li}

\affiliation[inst1]{organization={Institute of Biomedical Engineering, Chinese Academy of Medical Sciences \& Peking Union Medical College},
            addressline={236 Baidi Road, Nankai District}, 
            city={Tianjin},
            postcode={300192}, 
            state={Tianjin},
            country={China}}

\author[inst1]{Ziyue Ma}
\author[inst1]{Hong Wang\corref{cor1}}
\ead{wanghong@bme.pumc.edu.cn}
\author[inst1]{Juan Deng}
\author[inst1]{Yan Wang}
\author[inst1]{Zhenyu Xu}
\author[inst2]{Feng Yan}
\author[inst3]{Wenjun Tu}
\author[inst1]{Hong Sha}
\cortext[cor1]{Corresponding author: Hong Wang}

\affiliation[inst2]{organization={Department of Neurosurgery, Capital Medical University Xuanwu Hospital},
            addressline={45 Changchun Street, Xicheng District}, 
            city={Beijing},
            postcode={100053}, 
            state={Beijing},
            country={China}}

\affiliation[inst3]{organization={Institute of Radiation Medicine, Chinese Academy of Medical Sciences \& Peking Union Medical College},
            addressline={238 Baidi Road, Nankai District}, 
            city={Tianjin},
            postcode={300192}, 
            state={Tianjin},
            country={China}}

\begin{abstract}
Background and Objective: In neurosurgery, fusing clinical images and depth images that can improve the information and details is beneficial to surgery. We found that the registration of face depth images was invalid frequently using existing methods. To abundant traditional image methods with depth information, a method in registering with depth images and traditional clinical images was investigated. Methods: We used the dlib library, a C++ library that could be used in face recognition, and recognized the key points on faces from the structure light camera and CT image. The two key point clouds were registered for coarse registration by the ICP method. Fine registration was finished after coarse registration by the ICP method. Results: RMSE after coarse and fine registration is as low as 0.995913 mm. Compared with traditional methods, it also takes less time. Conclusions: The new method successfully registered the facial depth image from structure light images and CT with a low error, and that would be promising and efficient in clinical application of neurosurgery.
\end{abstract}




\end{frontmatter}


\section{Introduction}
In neurosurgery, using  CT or MRI images acquired before surgery, clinicians can determine the position of the lesion by navigation\cite{RN202}. The problems is how to obtain the relationship between the coordinate system of medical images and the coordinate system of patients during surgery\cite{RN216,RN217,RN218}. Supposing the brain tissue is a rigid body, extrinsic features like skin markers, anatomical features, and so on will be used in registration for acquiring the transfer matrix to transform from the coordinate system of traditional images to the coordinate system of the patient during the operation\cite{RN203}. Depth cameras become popular in recent years. We take the structure light camera for example. The infrared projector of the structure light camera can launch the light that was coded by grey code, laser fringe, sine fringe, etc\cite{RN211}. The light that is reflected will be received by the structure light camera to get the depth image. Depth images from the depth camera can improve the information and details in medical images after registering medical images\cite{RN212,RN213,RN214,RN215}. Above all, the feature from depth images is available for use in solving the coordinate system transformation after registering. It is more automated and easier to register with features from depth images rather than skin markers and so on. Therefore, there is great promise in neurosurgery by using depth cameras.

Depth images are widely used in surgery. Xiao et al\cite{RN204}. developed a deep learning model to estimate patient-specific reference bony shape models for patients with orthognathic deformities. It improves clinical workflows greatly. Lorsakul et al\cite{RN205}. reported a new point-cloud-to-point-cloud technique on tool calibration for dental implant surgical path tracking. It converges to the minimum error of 0.77\% with fewer post images of the tool. Hu et al\cite{RN206}. raised a general GAN architecture based on graph convolutional networks to generate 3D point clouds of brains from one single 2D image. The information from medical images will be improved by fusing medical images and these 3D point clouds. In recent years, the field of depth image registration has also been greatly developed. Koide et al\cite{RN207}. reported a new method called GICP to register 3D point clouds. It is faster than existing methods. Qin et al\cite{RN208}. developed a deep learning method called Geometric Transformer for robust superpoint matching. This method improves the inlier ratio by 17\%-30\% points and the registration recall by over 7 points. Yew et al\cite{RN209}. raised the 3DFeat-Net for 3D feature detectors and descriptors for point cloud matching using weak supervision. 3DFeat-Net obtains excellent performance on these gravity-aligned datasets. However, we found that the registration of face depth images is failed frequently with existing methods.

In this pursuit, the present study proposed a new facial depth image registration method that clinics can use. We used the dlib library, a C++ library that can be used in face recognition, to recognize the key points on faces from the structure light camera and CT image. The dlib library was first used in the medical field. We used the structure light camera (intel REALSENSE D415) to acquire the depth image of the patient’s face. The depth image is a part of the patient’s face point cloud. Depth image from CT image is extracted by VTK. We extracted the key points from depth images of the structure light camera and CT image. Then these two key points were registered for coarse registration by the ICP method. Fine registration was finished after coarse registration by the ICP method. We successfully registered the facial depth image from structure light images and CT with a low error. Our new method will increase the information and details that can be used in neurosurgery.

\section{Materials and Methods}

\subsection{Materials}

The structure light camera we used is intel REALSENSE D415. The version of the dlib library is 19.24. The version of the librealsense is 2.53.1. CT image was from a scan of one volunteer in this paper. The experiment was carried out on a laptop. CPU is intel Core i7-1165G7. The RAM size of the laptop is 16GB. The system of the laptop is Windows 10 22H2.
\subsection{Facial key points acquired}
The dlib library is a cross-platform open-source library. It was written in C++ programming language \cite{RN210}. Face recognition by deep learning is contained in the dlib library. The structure light camera can catch two video streams: RGB streaming and depth streaming. We each extract a frame from the two video streams in the meantime. The frame extracted from RGB streaming was called the RGB frame. The frame extracted from depth streaming was called depth frame. Face recognition in dlib was applied in the RGB frame. Then the facial key points from the face in the RGB frame were acquired. The facial key points from the depth frame were found by corresponding to the pixel from the RGB frame and depth frame. Through these key points, we segmented this depth image. Only the eyes and nose area were reserved because the patient may have a deformed mouth from a procedure like intubation.

The ITK library and the VTK library were used in acquiring the depth image from the CT image of a patient. The ITK library read the CT image of the patient. Marching Cubes, a method in the VTK library, was applied for extracting facial skin from a series of DICOM images that was read by the ITK library. The difficulty in acquiring facial key points from depth image from CT image is how to use the dlib library that recognizes from the 2D image mainly to recognize 3D depth image from CT image. The angle value of the normal vector from the face depth image acquired from the CT image was used in generating a 2D picture that was recognized by the dlib library. The height and width of this 2D picture were equal to the face-depth image. The value of each pixel was dependent on the angle value of the normal vector to the corresponding position of the face depth image. The dlib library recognized this 2D picture, and the facial key points of this 2D picture were acquired. The facial key points of the depth image from the CT image were also acquired by the coordinate of the facial key points from this 2D picture. Then we segmented the area that is the same as the depth image from the structure light camera manually.

\subsection{Coarse registration and fine registration}
The iterative closest point (ICP) algorithm is the dominant method in the registration of point clouds. But it cost a lot of time or leads to register failed in two point clouds that are big and far. Therefore, facial key points will be used in coarse registration by the ICP method. The source and target point clouds are closer than no coarse registration. Then fine registration will be finished by the ICP method. The transfer matrix will be acquired after fine registration. The outline of our registration method is shown in Figure 1.

\begin{figure}[htbp!]
\centering
\includegraphics[scale=0.8]{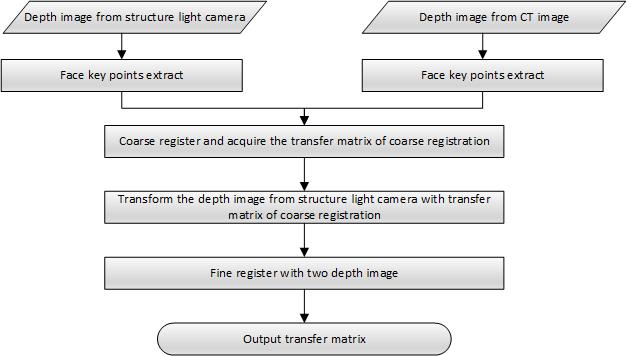}
\caption{The outline of the registration method.}
\label{fig:1}
\end{figure}

\section{Results}
\subsection{Depth images and facial key points acquired}
The depth images that we acquired are shown in Figure 2a. and Figure 2b. Figure 2a. is the depth image from the structure light camera. Figure 2b. is the depth image from the CT image. We only reserved the eyes and the nose area.

\begin{figure}[htbp!]
\subfigure[]{
\includegraphics{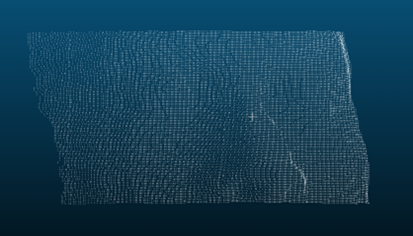}
\label{fig:2(a)}
}
\hspace{2mm}
\subfigure[]{
\includegraphics{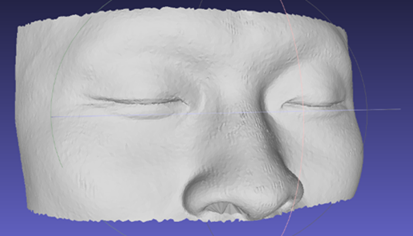}
\label{fig:2(b)}
}
\caption{The depth images acquired. (a) The depth image from the structure light camera. (b) The depth image from the CT image.}
\end{figure}

Figure 2b. was segmented for coarse registration and fine registration. The full face was used in generating a 2D face picture. Figure 3a. is a 2D face picture generated. Figure 3b. is the key points recognized by the dlib library.

\begin{figure}[htbp!]
\centering
\subfigure[]{
\includegraphics{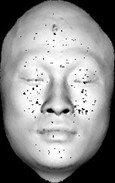}
\label{fig:3(a)}
}
\hspace{2mm}
\subfigure[]{
\includegraphics{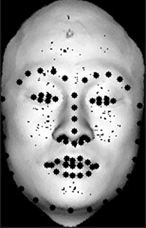}
\label{fig:3(b)}
}
\caption{The 2D face pictures. (a) The 2D face picture is generated from the depth image of the CT image. (b) The face key points are recognized by the dlib library.}
\end{figure}

In the facial key points of depth images, we reserved the key points of the eyes and the nose. Figure 4a. is the key points from the depth image of the structure light camera. Figure 4b. is the key points from the depth image of the CT image.

\begin{figure}[htbp!]
\subfigure[]{
\includegraphics{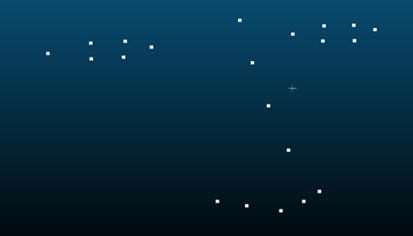}
\label{fig:4(a)}
}
\hspace{2mm}
\subfigure[]{
\includegraphics{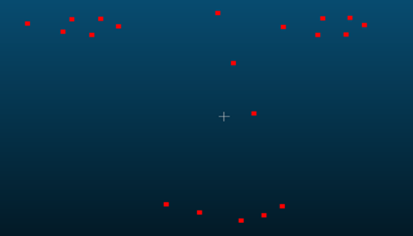}
\label{fig:4(b)}
}
\caption{The facial key points acquired. (a) The facial key points from the structure light camera. (b) The facial key points from the CT image.}
\end{figure}

\subsection{Coarse registration and fine registration}

The coarse registration was applied by the point cloud library (PCL). The coarse registration was finished by the ICP method. We set the maximum iterations as 200. Figure 5a. is the result of the facial key points registration. Figure 5b. is the result of coarse registration.

\begin{figure}[htbp!]
\subfigure[]{
\includegraphics{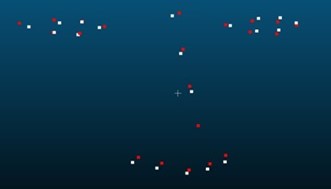}
\label{fig:5(a)}
}
\hspace{2mm}
\subfigure[]{
\includegraphics{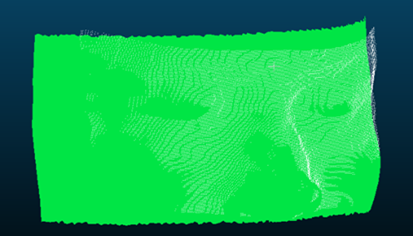}
\label{fig:5(b)}
}
\caption{The result of coarse registration. (a) The facial key points registration. The red points are from the structure light camera. The white points are from the CT image. (b) The result of coarse registration.}
\end{figure}

Figure 6. is the result of the fine registration. We used the software called CloudCompare to finish it. The method to register is the same as the coarse registration. We set the maximum iterations as 150. Final overlap was set as 100\%.

\begin{figure}[htbp!]
\centering
\includegraphics[]{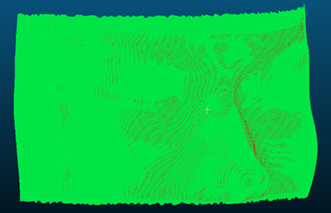}
\caption{The result of fine registration.}
\label{fig:6}
\end{figure}

\subsection{Accuracy and consuming}

The root means squared error (RMSE) is an indicator to access the accuracy of registration. We also acquired key points with 3 methods (ISS, Harris, and Sift). Coarse and fine registration was finished by the ICP method with the same parameters. Table 1. summarizes the result of the RMSE with our method and these 3 methods. The best results are denoted in bold font.

\begin{table}[ht]
\centering
\small
\begin{tabular}[b]{*{5}{p{0.8cm}<{\raggedright}}}
\multicolumn{5}{l}{\small{\textbf{Table 1}}}\\
\multicolumn{5}{l}{\small{The result of the RMSE with ours, ISS, Harris, and Sift methods.}}\\
\specialrule{0.05em}{3pt}{3pt}
\makebox[0.3\textwidth][c]{$Methods$}&\makebox[0.6\textwidth][c]{Ours}&\makebox[0.68\textwidth][c]{ISS}&\makebox[0.76\textwidth][c]{Harris}&\makebox[0.84\textwidth][c]{Sift}\\
\specialrule{0.05em}{2pt}{2pt}
\makebox[0.3\textwidth][c]{$Coarse\ registration (mm)$} & \makebox[0.6\textwidth][c]{\textbf{2.06693}}&\makebox[0.68\textwidth][c]{5.15501}&\makebox[0.76\textwidth][c]{5.05797}&\makebox[0.84\textwidth][c]{5.3156}\\
\makebox[0.3\textwidth][c]{$Fine\ registration (mm)$}&\makebox[0.6\textwidth][c]{0.995913}&\makebox[0.68\textwidth][c]{\textbf{0.995672}}&\makebox[0.76\textwidth][c]{4.98329}&\makebox[0.84\textwidth][c]{4.98339}\\
\specialrule{0.05em}{2pt}{0pt}
\end{tabular}
\end{table}
\normalsize

Figure 7. and Figure 8. show the result of registration with ISS, Harris, and Sift methods. Obviously, the methods with Harris and Sift had a failed registration.

\begin{figure}[htbp!]
\subfigure[]{
\includegraphics[scale=0.88]{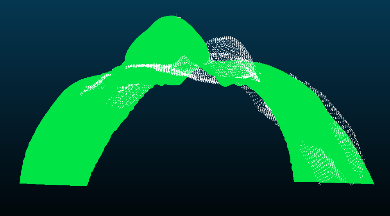}
\label{fig:7(a)}
}
\hspace{2mm}
\subfigure[]{
\includegraphics[scale=0.88]{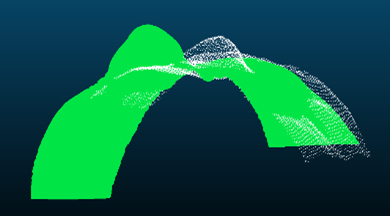}
\label{fig:7(b)}
}
\hspace{2mm}
\subfigure[]{
\includegraphics[scale=0.88]{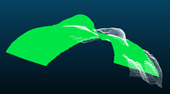}
\label{fig:7(c)}
}
\caption{The result of coarse registration. (a) ISS. (b) Harris. (c) Sift.}
\end{figure}

\begin{figure}[htbp!]
\subfigure[]{
\includegraphics[scale=0.88]{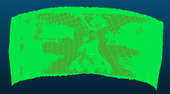}
\label{fig:8(a)}
}
\hspace{2mm}
\subfigure[]{
\includegraphics[scale=0.88]{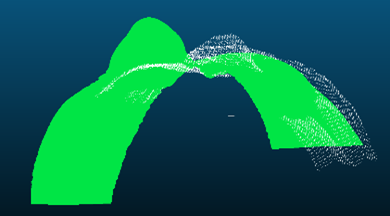}
\label{fig:8(b)}
}
\hspace{2mm}
\subfigure[]{
\includegraphics[scale=0.88]{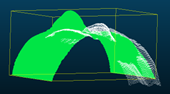}
\label{fig:8(c)}
}
\caption{The result of fine registration. (a) ISS. (b) Harris. (c) Sift.}
\end{figure}

We also assessed the cost time of our and ISS methods. Table 2. summarizes the result of the cost time with ours and the ISS method. $T_{c}$ is the cost time of coarse registration. $T_{f}$ is the cost time of fine registration. $T_{t}$ is the cost time of total. We tried each method three times. The average time we recorded in this table. The result is a four-place decimal. Obviously, our method costs less time as fewer features are used in coarse registration.

\begin{table}[ht]
\centering
\small
\begin{tabular}[b]{*{3}{p{0.8cm}<{\raggedright}}}
\multicolumn{3}{l}{\small{\textbf{Table 2}}}\\
\multicolumn{3}{l}{\small{The cost time of ours and ISS methods. $T_{c}$ is the cost time of coarse registration.}}\\
\multicolumn{3}{l}{\small{$T_{f}$ is the cost time of fine registration. $T_{t}$ is the cost time of total. The best results }}\\
\multicolumn{3}{l}{\small{are denoted in bold font.}}\\
\specialrule{0.05em}{3pt}{3pt}
\makebox[0.3\textwidth][c]{$Methods$}&\makebox[0.9\textwidth][c]{Ours}&\makebox[1.4\textwidth][c]{ISS}\\
\specialrule{0.05em}{2pt}{2pt}
\makebox[0.3\textwidth][c]{$Number\ of\ features$}&\makebox[0.9\textwidth][c]{21 vs. 20}&\makebox[1.4\textwidth][c]{421 vs. 489}\\
\makebox[0.3\textwidth][c]{$T_{c}(s)$}&\makebox[0.9\textwidth][c]{\textbf{0.0949}}&\makebox[1.4\textwidth][c]{0.8640}\\
\makebox[0.3\textwidth][c]{$T_{f}(s)$}&\makebox[0.9\textwidth][c]{\textbf{25.2648}}&\makebox[1.4\textwidth][c]{26.2925}\\
\makebox[0.3\textwidth][c]{$T_{t}(s)$}&\makebox[0.9\textwidth][c]{\textbf{25.3598}}&\makebox[1.4\textwidth][c]{27.1565}\\
\specialrule{0.05em}{2pt}{0pt}
\end{tabular}
\end{table}
\normalsize

\section{Discussion}
We found that the registration of face depth images is failed frequently with existing methods. In this pursuit, the present study envisaged a new facial depth image registration method that clinics can use. We used the dlib library to recognize the key points from depth images from the structure light image and the CT image. The error of registration is lower, and the registration costs less time. In our experiment, the root means squared error (RMSE) is 0.995913 mm. However, we deem that the RMSE of registration can be lower than now. The volume of the structure light camera we used is small. The button-hole space of the light camera we used is short. The longer button hole space is, the higher accuracy of depth image we can get. We think that more accurate structured light camera can be used to lower the value of RMSE. We made a library called libFaceRegistration with the codes in this pursuit after the experiment. The library is available at https://github.com/binfenseca2969/libFaceRegistration.

\section{Conclusions}
We successfully registered the facial depth image from structure light images and CT with a low error. The dlib library was first used in the medical registration field. We used the structure light camera (intel REALSENSE D415) to acquire the depth image of the patient’s face. The depth image is a part of the patient’s face point cloud. Depth image from CT image is extracted by VTK. We extracted the key points from depth images of the structure light camera and CT image by the dlib library. Then these two key points were registered for coarse registration by the ICP method. Fine registration was finished after coarse registration by the ICP method. The issue that registration of face depth images is failed frequently will be resolved. Our new method will increase the information that can be used in neurosurgery. The clinicians can transfer the coordinate system from the traditional clinic images like CT to the patient with our method. More convenience will be acquired for the surgical robot in neurosurgery.

\section{Acknowledgments}
This study was supported by grants from the CAMS Innovation Fund for Medical Science (No. 2021-I2M-1-042, 2021-I2M-1-015).



 \bibliographystyle{elsarticle-num} 
 \bibliography{library}





\end{document}